\documentclass[letterpaper]{article} 
\usepackage[submission]{aaai25}  
\usepackage{times}  
\usepackage{helvet}  
\usepackage{courier}  
\usepackage[hyphens]{url}  
\usepackage{graphicx} 
\usepackage{natbib}  
\usepackage{caption} 
\usepackage{algorithm}
\usepackage{amsmath}
\usepackage{amssymb}
\usepackage{cleveref}
\usepackage{algorithmic}
\usepackage{booktabs}
\usepackage{tcolorbox}
\usepackage{colortbl}
\usepackage{dashrule}
\definecolor{mygray-bg}{gray}{0.9}
\newcommand{\eg}{e.g.,}

\usepackage{newfloat}
\usepackage{listings}
\DeclareCaptionStyle{ruled}{labelfont=normalfont,labelsep=colon,strut=off} 
\floatstyle{ruled}
\newfloat{listing}{tb}{lst}{}
\floatname{listing}{Listing}
\title{Empowering LLMs with Pseudo-Untrimmed Videos for Audio-Visual\\Temporal Understanding}
\author{%
  \textbf{
      Yunlong Tang\textsuperscript{1},
      Daiki Shimada\textsuperscript{2},
      Jing Bi\textsuperscript{1},
      Mingqian Feng\textsuperscript{1},\\
      Hang Hua\textsuperscript{1},
      Chenliang Xu\textsuperscript{1,}\thanks{Corresponding author.}
  }
}
\affiliations{
\textsuperscript{1}University of Rochester, \textsuperscript{2}Sony Group Corporation
  \\
  {\tt 
  \{yunlong.tang, jing.bi, mingqian.feng, chenliang.xu\}@rochester.edu, 
  }\\
  {\tt 
   hhua2@cs.rochester.edu, Daiki.Shimada@sony.com
  }\\
}

\usepackage{bibentry}

\begin{document}

\maketitle

\begin{abstract}

Large language models (LLMs) have demonstrated remarkable capabilities in natural language and multimodal domains. By fine-tuning multimodal LLMs with temporal annotations from well-annotated datasets, e.g., dense video captioning datasets, their temporal understanding capacity in video-language tasks can be obtained. However, there is a notable lack of untrimmed audio-visual video datasets with precise temporal annotations for events. This deficiency hinders LLMs from learning the alignment between time, audio-visual events, and text tokens, thus impairing their ability to localize audio-visual events in videos temporally.
To address this gap, we introduce PU-VALOR, a comprehensive audio-visual dataset comprising over 114,000 pseudo-untrimmed videos with detailed temporal annotations. PU-VALOR is derived from the large-scale but coarse-annotated audio-visual dataset VALOR, through a subtle method involving event-based video clustering, random temporal scaling, and permutation.
By fine-tuning a multimodal LLM on PU-VALOR, we developed AVicuna, a model capable of aligning audio-visual events with temporal intervals and corresponding text tokens. AVicuna excels in temporal localization and time-aware dialogue capabilities.
Our experiments demonstrate that AVicuna effectively handles temporal understanding in audio-visual videos and achieves state-of-the-art performance on open-ended video QA, audio-visual QA, and audio-visual event dense localization tasks.
\end{abstract}

%

\section{Introduction}
\label{sec:intro}

\begin{figure*}[!ht]
    \centering
\includegraphics[width=0.95\linewidth]{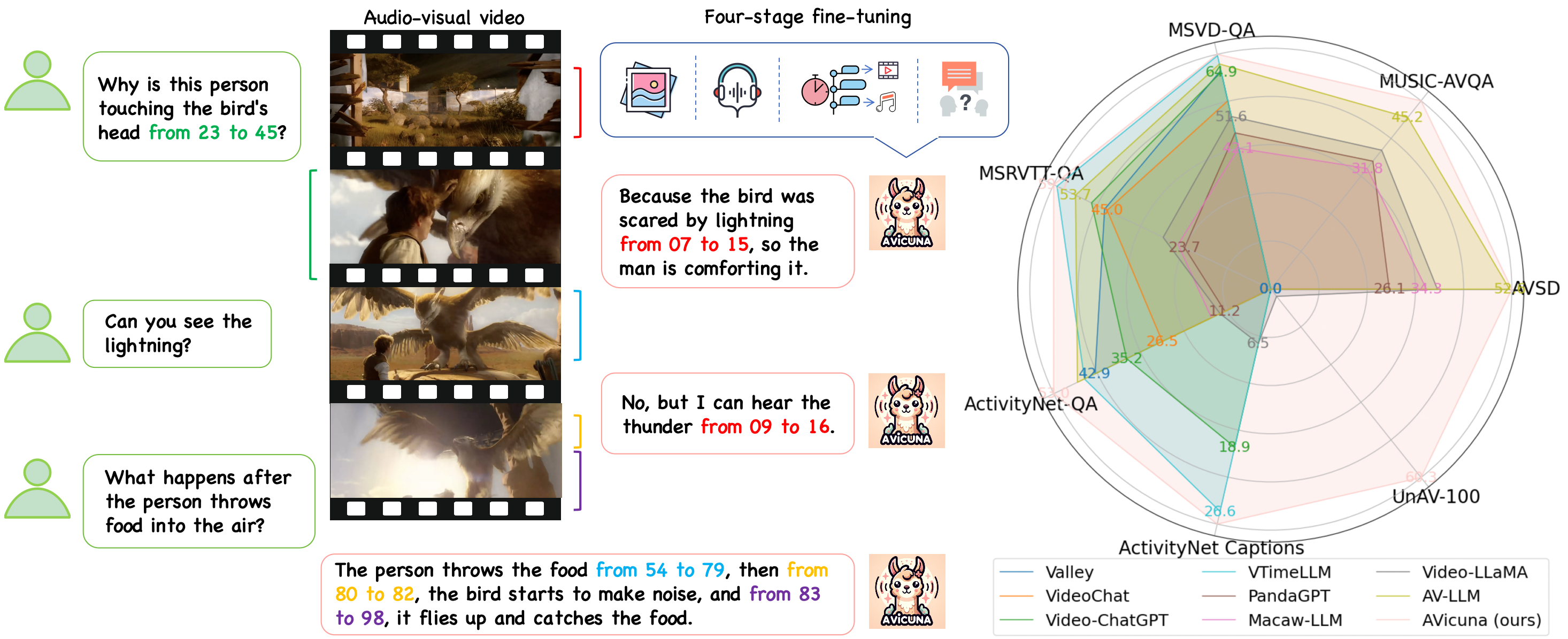}
    \caption{\textbf{Left}: AVicuna's four-stage fine-tuning aligns natural language with exact time segments in audio-visual videos, highlighting its adeptness in dynamic content analysis. \textbf{Right}: AVicuna's superior performance across various video and audio-visual understanding tasks compared to other models.}
    \label{fig:teaser}
    \vspace{-1.0em}
\end{figure*}

Large Language Models (LLMs) have recently advanced natural language processing (NLP), evolving into Multimodal LLMs (MLLMs) capable of comprehending various modalities like text, images, audio, and videos~\cite{chen2023shikra, videollama,llava,video-llava,xu2023launchpadgpt,zhang2023dnagpt}. 
Despite the advancements, MLLMs still struggle to provide a fine-grained understanding of spatial or temporal details in multimodal contexts. To address these limitations, several works have explored ~\cite{chen2023shikra, xuan2023pink} incorporating object bounding box coordinates in natural language format into image-text data, enabling MLLMs to directly identify the location of objects in images using natural language and respond accurately to user-provided bounding box coordinates, thereby enhancing fine-grained region-level understanding~\cite{chen2023shikra, xuan2023pink, peng2023kosmos, bai2023qwen}.
While MLLMs have demonstrated potential in fine-grained image understanding, it's equally crucial to extend their capabilities to the video domain to achieve detailed video comprehension.
Some recent advancements~\cite{yang2023vid2seq,li2023videochat,wang2023chatvideo,huang2023vtimellm,zhang2024cocot} have leveraged natural language for temporal predictions in video understanding tasks, such as dense video captioning, video temporal grounding, etc. These approaches have demonstrated competitive performance compared to traditional regression-based methods while retaining general capabilities, such as video question answering (Video QA). 
However, these methods predominantly concentrate solely on visual content, overlooking the dynamic and multimodal nature of the real world. 
For example, audio-visual content in videos represents a common form of such data. By integrating various modalities, including sound, we can achieve a more comprehensive analysis of content.
As we delve deeper into this integration, it faces two significant challenges:

(1) In contrast to the abundance of dense video caption datasets, the audio-visual domain faces a significant bottleneck due to the lack of datasets providing detailed audio-visual event captions with accurate timestamp annotations.

(2) Developing audio-visual learning methods that effectively capture the intricate blend of auditory and visual cues, enabling them to interpret complex, intertwined information across various events within videos.

To tackle the challenge (1), we propose a practical yet straightforward pipeline that leverages the VALOR-32K~\cite{chen2023valor} dataset with high-quality audio-visual captions to construct  PU~(\textbf{P}seudo-\textbf{U}ntrimmed)~-VALOR dataset contains audio-visual videos with corresponding temporal boundary annotations. 
The PU-VALOR dataset is created by applying Random Temporal Scaling and Permutation to videos clustered by captions. This innovative approach, aimed at generating Pseudo-Untrimmed videos, theoretically enables the creation of an unlimited number of untrimmed videos. 
Consequently, the PU-VALOR dataset features over 114k video-caption pairs, each annotated with precise temporal boundaries, thus offering a contribution toward enriching the audio-visual research domain.

Moving forward from the proposed dataset, we recognize the second challenge in audio-visual learning: accurately modeling the temporal dynamics in untrimmed audio-visual content. 
Previous methods~\cite{avllm,videollama,pandagpt,macawllm} have often combined embeddings from different modalities into single embeddings without adequately considering their intrinsic temporal relation.
To tackle this critical issue, we introduce AVicuna, which comprises Multimodal Encoders, two Connective Adapters, an Audio-Visual Token Interleaver (AVTI), and an LLM. Multimodal Encoders extract embeddings from vision and audio modalities, which are aligned with the LLM's token space through Connective Adapters. 
The AVTI orchestrates the temporal relation of tokens from audio and video by creating interleaved audio-visual token sequences as inputs for the LLM.
We employ a multi-stage fine-tuning approach to enhance AVicuna's capabilities, focusing on four critical stages: Vision-Text Alignment, Audio-Text Alignment, Time-Event Alignment, and Instruction Tuning. To foster effective alignment between multimodal tokens and LLM's token space, we have also aggregated several audio datasets, including AudioSet~\cite{audioset}, AudioCap~\cite{kim-etal-2019-audiocaps}, and Auto-ACD~\cite{autocad}, to form a comprehensive audio-text dataset with 222K pairs, termed A5-222K (\textbf{A}udio-text \textbf{A}lignment with \textbf{A}udioSet, \textbf{A}udioCap, and \textbf{A}uto-CAD).

Our experiments demonstrate that the AVicuna fine-tuned on PU-VALOR achieves outstanding performance in both coarse-grained QA tasks and fine-grained temporal understanding tasks, as \Cref{fig:teaser} shown. It surpasses most LLM-based video understanding models and sets a new benchmark in the Audio-Visual Event Dense Localization (AVEDL) task.

In summary, our contributions are three-fold:
\begin{itemize}
    \item We propose a novel approach to synthesize pseudo-untrimmed audio-visual videos and corresponding temporal boundary annotations using high-quality captions from the VALOR dataset, resulting in the PU-VALOR dataset.
    \item We introduce AVicuna, an audio-visual LLM with an Audio-Visual Token Interleaver and Time-Event Alignment Tuning on the PU-VALOR dataset, which achieves temporal synchronism and fine-grained understanding in audio-visual videos.
    \item Our experiments demonstrate that AVicuna significantly advances the state-of-the-art in the AVEDL task and exhibits strong performance in both coarse-grained QA and fine-grained temporal understanding tasks.
\end{itemize}

\section{Related Work}
\begin{figure*}[!ht]
    \centering
    \includegraphics[width=\linewidth]{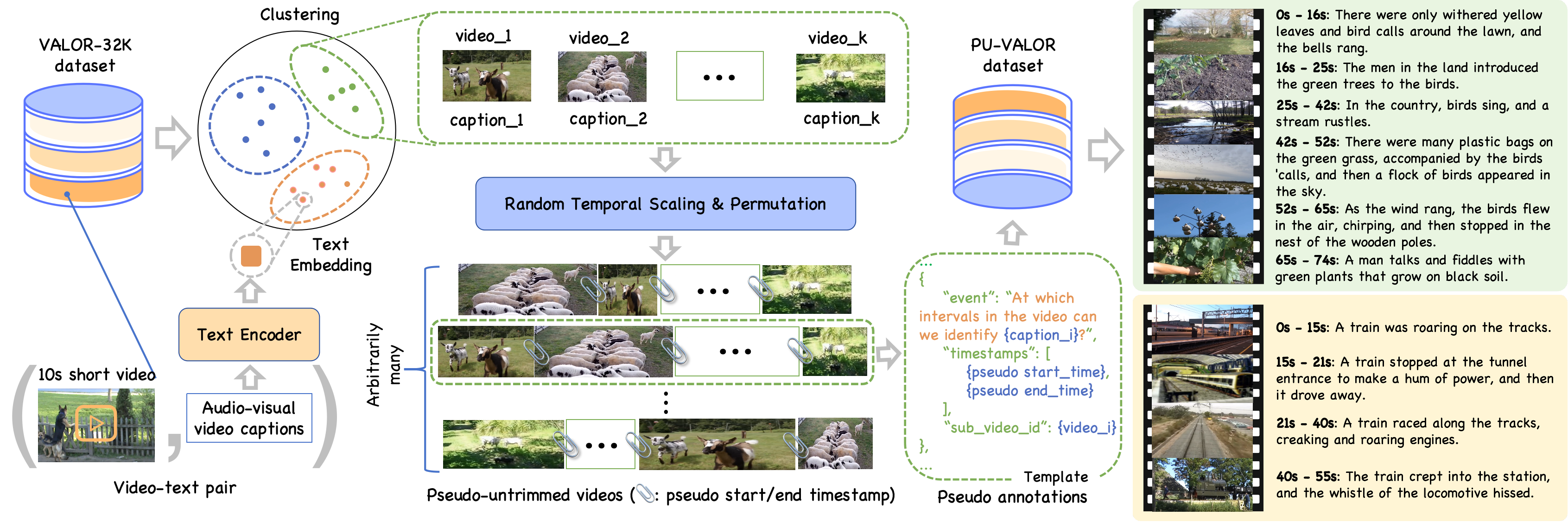}
    \caption{Pipeline for creating the PU-VALOR dataset, which involves extracting text embeddings from high-quality audio-visual captions of the original trimmed VALOR-32K dataset, clustering these embeddings, and then applying Random Temporal Scaling \& Permutation to generate pseudo-untrimmed videos. These synthesized videos are then annotated with temporal boundaries using a template-based approach to facilitate the following audio-visual time-event alignment.}
    \label{fig:pipline}
    \vspace{-1.0em}
\end{figure*}
\subsubsection{Untrimmed Video Understanding.}
Temporal localization is key for understanding long-form videos by linking specific segments to their semantics. Key tasks include temporal video grounding~\cite{luo2023towards,wang2022negative}, dense video captioning~\cite{wang2021end,yang2023vid2seq}, and video highlight detection~\cite{lei2021detecting,jiang2024single}.
However, action localization and highlight detection models often rely on predefined labels~\cite{zhang2022actionformer}, limiting the scope. Recent work on event boundary detection~\cite{shou2021gebd,wang2022geb+,tang2023llmva} and dense captioning~\cite{anet_cap} aims to address these constraints using event description datasets~\cite{lin2023videoxum}. Despite this, most models still rely on regression for temporal predictions, requiring extra heads for captioning and regression~\cite{wang2021pdvc,zhang2022contextgebc,tang2022multi}. Recent LLM advancements offer a shift by using natural language to directly specify temporal locations, offering a more intuitive approach.

\subsubsection{Multimodal LLMs.} The development of Multimodal LLMs (MLLMs) has been driven by advancements in LLMs, enabling the integration of multimodal inputs~\cite{alayrac2022flamingo,llava,hu2022promptcap,wang2023caption,hua2024finematch,lyu2023gpt,bi2023misar}. {Flamingo}~\cite{alayrac2022flamingo} utilizes visual in-context learning for visual question answering, while {LLaVA}~\cite{llava} introduces visual instruction tuning to enhance visual understanding. Models like {VisionLLM}~\cite{wang2024visionllm}, {KOSMOS-2}~\cite{peng2023kosmos}, ~\cite{chen2023shikra}, and {Qwen-VL}~\cite{bai2023qwen} further advance MLLMs with visual grounding. Recent advancements such as {VideoChat}~\cite{li2023videochat}, {ChatVideo}~\cite{wang2023chatvideo}, {V2Xum-LLM}~\cite{hua2024v2xum}, and {VTimeLLM}~\cite{huang2023vtimellm} extend this fine-grained understanding to dynamic video content using natural language to define temporal boundaries without special tokens. We integrate audio cues to offer a more comprehensive approach to audio-visual temporal understanding tasks.

\subsubsection{Audio-Visual Video Datasets.}
Increasing attention is being directed toward LLMs that support audio-visual inputs, such as Video-LLaMA~\cite{videollama}, PandaGPT~\cite{pandagpt}, Macaw-LLM~\cite{macawllm}, and AV-LLM~\cite{avllm}, which are trained on audio-visual video datasets to enhance understanding of audio-visual content. However, these models struggle with fine-grained understanding of long or untrimmed videos due to the lack of detailed annotations in existing datasets. The VALOR dataset~\cite{chen2023valor} offers high-quality audio-visual captions but consists of trimmed 10-second clips. Other datasets like VGG-Sound-AVEL100K~\cite{VggSound-AVEL100k}, AVVP~\cite{tian2020AVVP}, UnAV-100~\cite{unav100}, and LFAV~\cite{lfav} provide temporal annotations but lack rich captioning. While AVSD~\cite{avsd} and MUSIC-AVQA~\cite{music-avqa} offer quality question-answer pairs, their temporal questions lack precise timestamps. This gap in datasets limits the models' ability to learn the relationship between audio-visual context and temporal boundaries. \Cref{tab:data_comp} compares important features across different video datasets, including untrimmed videos, audio-visual modalities, captions, and timestamps. 
\begin{table}[!ht]
\centering
\resizebox{\columnwidth}{!}{%
\setlength{\tabcolsep}{4pt}
\begin{tabular}{l|cccc}
\toprule
\textbf{Dataset} & \textbf{\begin{tabular}[c]{@{}c@{}}Un-\\ trimmed\end{tabular}} & \textbf{\begin{tabular}[c]{@{}c@{}}Audio-\\ Visual\end{tabular}} & \textbf{Captions} & \textbf{\begin{tabular}[c]{@{}c@{}}Time-\\ stamps\end{tabular}} \\ \midrule\midrule
ActNetCaps~\cite{anet_cap} & $\checkmark$ & $\times$ & $\checkmark$ & $\checkmark$ \\
InternVid~\cite{wang2023internvid} & $\checkmark$ & $\times$ & $\checkmark$ & $\checkmark$ \\
VGG-Sound-AVEL100K & $\times$ & $\checkmark$ & $\times$ & $\checkmark$ \\
AVVP~\cite{tian2020AVVP} & $\times$ & $\checkmark$ & $\times$ & $\checkmark$ \\
LFAV~\cite{lfav} & $\checkmark$ & $\checkmark$ & $\times$ & $\checkmark$ \\
UnAV-100~\cite{unav100} & $\checkmark$ & $\checkmark$ & $\times$ & $\checkmark$ \\
VALOR~\cite{chen2023valor} & $\times$ & $\checkmark$ & $\checkmark$ & $\times$ \\
\rowcolor[HTML]{ECF4FF} PU-VALOR (ours) & $\checkmark$ & $\checkmark$ & $\checkmark$ & $\checkmark$ \\ \bottomrule
\end{tabular}%
}
\caption{Comparison of datasets based on untrimmed videos, audio-visual modalities, captions, and timestamps, showcasing the full coverage of all attributes by our PU-VALOR dataset.}
\label{tab:data_comp}
\vspace{-1.0em}
\end{table}

\section{Methodology}


\subsection{PU-VALOR Dataset}
\label{sec:data}
One of the primary challenges in untrimmed audio-visual video understanding is the scarcity of datasets with fine-grained annotations for temporal audio-visual events. To tackle this issue, we propose a practical yet straightforward pipeline, as illustrated in \Cref{fig:pipline}, to utilize the existing VALOR-32K audio-visual dataset~\cite{chen2023valor}, which comprises exclusively trimmed videos.
 By synthesizing untrimmed videos with precise temporal labels, we have created the PU-VALOR dataset, that enable  LLMs from learning the alignment between time, audio-visual events, and text token.


\subsubsection{Clustering Videos with Similar Events.}
When creating untrimmed videos from trimmed clips, it’s crucial to maintain semantic coherence within the untrimmed video to ensure a natural flow of content. This means that transitions between different segments should not be abrupt or disjointed, as such sudden shifts can disrupt the viewer’s understanding.
To ensure the content within an untrimmed video is semantically related, we group similar video segments based on the semantic similarity of their captions for follow-up untrimmed video generation. 

We utilize a text encoder $ E_{\text{txt}} $ to embed captions from the video-caption pairs \(\{(v_1, c_1), (v_2, c_2), \ldots, (v_n, c_n)\}\) sourced from the VALOR-32K dataset.
For each caption \( c_i \), the embedding is given by: $\bar{c}_i = E_{\text{txt}}(c_i)$
Next, we apply a clustering algorithm to the set of embeddings $ \{\bar{c}_1, \bar{c}_2, \ldots, \bar{c}_n\} $ to identify and group videos with similar events, resulting in clusters denoted as  $ K = \{k_1, k_2, \ldots, k_m\} $.
From each video cluster $k$, we random select one and then identify the $m-1$ top most similar clips within the same cluster. This selection is defined by:
\begin{equation}
    S_k = \{v_i \mid v_i \in V_k, \text{ where } i \in \{1, 2, \ldots, m\}\}.
\end{equation}
We repeat this process for each cluster until no more than $m$ clips remain in any cluster.
This process ensures that each set of $m$ clips is closely related in content.

\subsubsection{Random Temporal Scaling \& Permutation.}
To ensure diverse temporal relationships between the clips, for each selected video $v_i \in S_k$, we randomly scale its duration within the range $[T_{\text{min}}, T_{\text{max}}]$ as
\begin{equation}
    T_{v_i}^{\text{new}} = T_{v_i} \times \text{random}(T_{\text{min}},~T_{\text{max}}).
\end{equation}
Then, we shuffle the order of the selected videos $S_k$ and concatenate the videos in $S_k$ to form a new untrimmed video $U_k$:
\begin{equation}
    U_k = v_{\pi(1)} || v_{\pi(2)} || \ldots || v_{\pi(m)},
\end{equation}
   where $\pi$ is a random permutation of the indices $\{1, 2, \ldots, m\}$.

\subsubsection{Annotation.}
As we know the original duration $ T_{v_i} $ and the scaled duration $ T_{v_i}^{\text{new}} $ for each video, we can map captions to specific temporal intervals in $ U_k $. If the original duration for a caption in $ v_i $ is $ d_i $, the new timestamp in $ U_k $ becomes:
\begin{equation}
    [t_{\text{start},i}^{\text{new}},~t_{\text{end},i}^{\text{new}}] = [T_{\text{offset,i}},~T_{\text{offset,i}} + d_i \cdot \epsilon]
\end{equation}
   where $ T_{\text{offset}} $ is the cumulative duration of all preceding videos in $ U_k $ after scaling, and the scaling factor $\epsilon= T_{v_i}^{\text{new}} / T_{v_i} $.
We annotate temporal intervals $[t_{\text{start},i}^{\text{new}}, t_{\text{end},i}^{\text{new}}]$ within $ U_k $ to correspond with the content described by the captions $c_i$.

\subsubsection{Data Visualization.}
Some pseudo-untrimmed video examples from the PU-VALOR constructed are shown in \Cref{fig:pipline}: (Top) The sub-videos that make up this pseudo-untrimmed video share common themes of nature, including elements such as birds, greenery, and natural sounds, set against a backdrop of changing environments and human interaction. (Bottom) the sub-videos in this example all revolve around trains, capturing their movements on the tracks, associated sounds, and interactions with the environment, such as entering tunnels and stations. Clustering videos with similar captions or events reduces the dramatic changes in semantics and content, which avoids the model learning a shortcut to localize the temporal boundaries.


\subsection{AVicuna Model}

\begin{figure*}[!ht]
    \centering
    \includegraphics[width=0.95\linewidth]{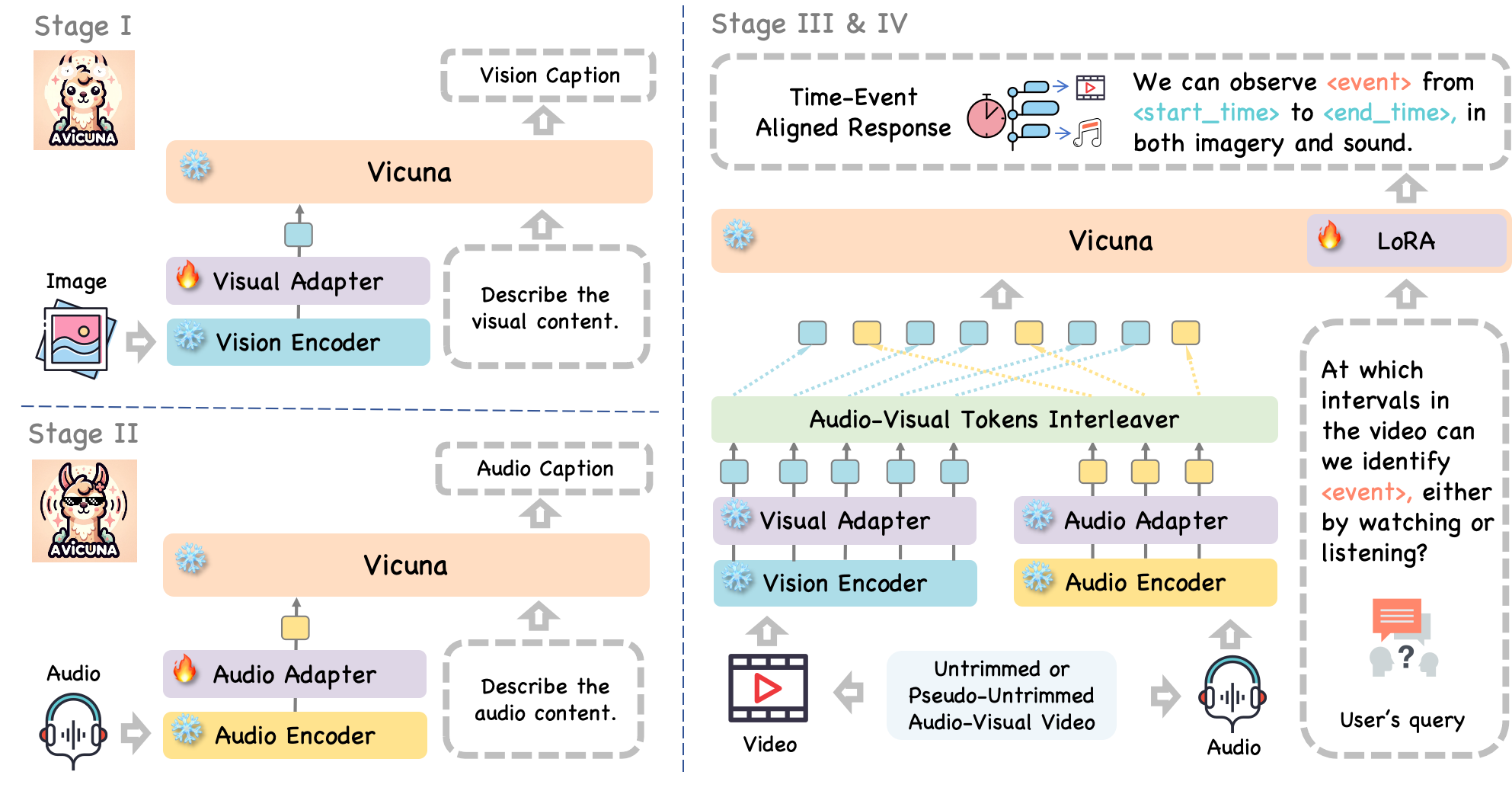}
    \caption{AVicuna model architecture and fine-tuning process. Vision and Audio Adapters are MLPs that align modalities with LLM. The Audio-Visual Tokens Interleaver ensures temporal synchronization. LoRA fine-tuning aligns temporal boundaries with events and enhances instruction-following capabilities.}
    \label{fig:model}
    \vspace{-1.0em}
\end{figure*}
\subsubsection{Overview.}
\Cref{fig:model} illustrates AVicuna's architecture, comprising Multimodal Encoders, Connective Adapters, an Audio-Visual Token Interleaver (AVTI), and an LLM. The encoders extract embeddings aligned to the LLM's token space, and the AVTI interleaves them, with an Audio-Interleaving Rate (AIR) enhancing temporal synchronism.

\subsubsection{Multimodal Encoders.}
Multimodal Encoder includes Vision Encoder and Audio Encoder. For visual input, we employ the CLIP ViT-14/L~\cite{clip} as Vision Encoder to extract visual embeddings $F=\{f_i\}_{i=1}^M$, where $M$ denotes the number of visual embeddings. When the input is image, $M=1$. For audio input, we utilize CLAP~\cite{elizalde2023clap} as Audio Encoder to obtain audio embeddings $A=\{a_i\}_{i=1}^N$, with $N$ representing the number of audio embeddings.

\subsubsection{Connective Adapters.} To avoid interference between the different modalities during the alignment, we adopt two MLPs as Vision Adapter and Audio Adapter for visual embeddings and audio embeddings, respectively, to get visual tokens $\bar{F}=\{\bar{f}_i\}_{i=1}^M$ and audio tokens $\bar{A}=\{\bar{a}_i\}_{i=1}^N$ that are aligned with LLM's token space.

\subsubsection{Audio-Visual Tokens Interleaver.}
Unlike simply adding positional embeddings to audio and video embeddings in previous work~\cite{macawllm,avllm}, the Audio-Visual Token Interleaver (AVTI) rearranges video and audio embeddings without altering their sequence order, keeping the overall token length constant. The audio-interleaving rate (AIR), denoted as $\rho\in[0,1]$, controls the ratio of video to audio tokens. Audio tokens and video tokens are interpolated or downsampled to $\tilde{A}=\{\tilde{a}_{i}\}_{i=1}^{\tilde{N}}$ and $\tilde{F}=\{\tilde{f}_{i}\}_{i=1}^{\tilde{M}}$, where $\tilde{N}=\rho T$ and $\tilde{M}=(1-\rho)T$, with $T$ being the total sequence length generated by AVTI. As shown in \Cref{fig:model}, AVTI systematically interleaves audio and video tokens, preserving their original order to maintain temporal alignment. The output, represented by the audio-visual context $\Psi=\{\psi_t\}_{t=1}^T$, is given by:
\begin{equation}
     \psi_t = 
\begin{cases} 
 \tilde{a}_{\lceil t / (\omega_\rho + 1) \rceil} & \text{if}~ t\hspace{-7pt}\mod(\omega_\rho + 1) \equiv 0, \\
\tilde{f}_{\lceil t / (\omega_\rho + 1) \rceil} & \text{otherwise},
\end{cases}
\end{equation}
where $\omega_\rho=\lfloor{\frac{1-\rho}{\rho}}\rfloor$. Each context token represents both the audio-visual content and its temporal position within the video.

\subsubsection{Large Language Model.} We use the fine-tuned Vicuna-7B-v1.5~\cite{touvron2023llama} as our LLM to process interleaved audio-visual tokens and user queries $Q$, generating responses $R$:
\begin{equation}
R=\text{LLM}(I,Q),
\label{eq
}
\end{equation}where $I \in \{\bar{F},\bar{A},\Psi\}$ represents vision, audio, or both.

\subsection{Multi-stage Fine-tuning}
\label{sec:multi-stage-fine-tuning}
In AVicuna training, aligning embeddings from other modalities to the LLM's token space is preferred, using Video-Text and Audio-Text Alignment. To enable specific temporal patterns, we fine-tune LoRA~\cite{lora} during Time-Event Alignment. Recent studies~\cite{huang2023vtimellm} indicate this may affect question-answering, mitigated by Instruction Tuning. We thus employ a four-stage fine-tuning process, with datasets detailed in the Appendix.

\subsubsection{Stage I \& II: Multimodal-Text Alignment.}
In Vision-Text Alignment, we fix the Vision Encoder and LLM, updating only the Vision Adapter, using LCS-558K~\cite{llava}, a 558K image-text pair subset from LAION-CC-SBU with BLIP~\cite{li2023blip} captions. Here, $I=\bar{F}$, $Q$ is a visual content query, and $R$ is the image caption. In Audio-Text Alignment, we fix the Audio Encoder and LLM, updating only the Audio Adapter, using A5-222K, a 222K audio-text dataset compiled from AudioSet~\cite{audioset}, AudioCap~\cite{kim-etal-2019-audiocaps}, and Auto-ACD~\cite{autocad}. Here, $I=\bar{A}$, $Q$ is an audio content query, and $R$ is the audio caption.

\subsubsection{Stage III: Time-Event Alignment.} We freeze the fine-tuned Audio and Vision Adapters, updating only the LoRA~\cite{lora} parameters in the LLM. We create $Q$-$R$ pairs with time-related information from the PU-VALOR datasets, including (1) Time-referenced Query, Time-agnostic Response: $\{(Q,R)|\tau \sqsubseteq Q, \tau \not\sqsubseteq R)\}$, and (2) Time-agnostic Query, Time-referenced Response: $\{(Q,R)|\tau \not\sqsubseteq Q, \tau \sqsubseteq R)\}$, where $\tau:=$"\textit{from}~${\tau_s}$~\textit{to}~${\tau_e}$" and $\tau_s, \tau_e$ are event time points. Inputs ($I$) can be visual, audio, or combined; we use the InternVid~\cite{wang2023internvid} dataset to enrich visual event alignment training.

\begin{figure}[!ht] 
\centering
\begin{tcolorbox}[colback=white, colframe=gray, text width=0.85\columnwidth, title={\small Prompts for Time-Event Alignment.}, fontupper=\small, fontlower=\small]
$Q_1$: Tell me about the visual and audio events $\tau$ in the video.\\
$Q_2$: What was going on visually and audibly $\tau$ in the video?\\
$Q_3$: Please recount what occurred, including both video and audio, $\tau$ in the video.\\
$Q_4$: Could you tell me what happened, in terms of both imagery and sound, $\tau$ in the video?\\
$Q_5$: Provide details about the visual scenes and audio events $\tau$ in the video.\\
$Q_6$: Can you describe what occurred, both visually and audibly, $\tau$ in the video?\\
$Q_7$: Explain what happened, considering both video and audio, $\tau$ in the video.
\end{tcolorbox}
\vspace{-2em}
\end{figure}

\subsubsection{Stage IV: Instruction Tuning.}
Finally, we fine-tune AVicuna on instruction-following datasets, including UnAV-100~\cite{unav100} for event localization, and other instruction datasets like VideoInstruct100K~\cite{video-chatgpt}, ActivityNet Captions~\cite{anet_cap}, and DiDeMo~\cite{didemo}, to enhance question-answering and mitigate previous tuning effects~\cite{huang2023vtimellm}. At this stage, $I$ is also $\Psi$ or $\bar{F}$, $Q$ is a general instruction, and $R$ is the corresponding response.

\section{Experimental Results}
\begin{table*}[!ht]
\centering
\resizebox{\textwidth}{!}{%
\setlength{\tabcolsep}{4pt}
\begin{tabular}{l|cccc|cc|ccc}
\toprule
\textbf{Method} & \textbf{A\&V} & \textbf{TU} & \textbf{\#Pairs} & \textbf{LLM-size} & \textbf{AVSD} & \textbf{MUSIC-QA} & \textbf{MSVD-QA} & \textbf{MSRVTT-QA} & \textbf{ActivityNet-QA} \\ \midrule\midrule
Valley~\cite{luo2023valley} & $\times$ & $\times$ & 1.5M & 13B & - & - & 65.4 & 45.7 & 26.5 \\
VideoChat~\cite{li2023videochat} & $\times$ & $\checkmark$ & 25M & 7B & - & - & 56.3 & 45.0 & 26.5 \\
Video-ChatGPT~\cite{video-chatgpt} & $\times$ & $\checkmark$ & 0.9M & 7B & - & - & 64.9 & 49.3 & 35.2 \\
VTimeLLM~\cite{huang2023vtimellm} & $\times$ & $\checkmark$ & 0.7M & 7B & - & - & 69.8 & 58.8 & 45.5 \\
PandaGPT~\cite{pandagpt} & $\checkmark$ & $\times$ & 128M & 13B & 26.1 & 33.7 & 46.7 & 23.7 & 11.2 \\
Macaw-LLM~\cite{macawllm} & $\checkmark$ & $\times$ & 0.3M & 7B & 34.3 & 31.8 & 42.1 & 25.5 & 14.5 \\
AV-LLM~\cite{avllm} & $\checkmark$ & $\times$ & 1.6M & 13B & 52.6 & 45.2 & 67.3 & 53.7 & 47.2 \\
Video-LLaMA~\cite{videollama} & $\checkmark$ & $\checkmark$ & 2.8M & 7B & 36.7 & 36.6 & 51.6 & 29.6 & 12.4 \\
\rowcolor[HTML]{ECF4FF}{AVicuna (ours)} & $\checkmark$ & $\checkmark$ & 1.1M & 7B & \textbf{53.1} & \textbf{49.6} & \textbf{70.2} & \textbf{59.7} & \textbf{53.0} \\ \bottomrule
\end{tabular}%
}
\caption{Comparison with existing LLM-based methods on open-ended video QA (MSVD-QA, MSRVTT-QA, ActivityNet-QA) and AVQA (AVSD, MUSIC-AVQA) benchmarks. \textbf{A\&V}: the model supports both video and audio input. \textbf{TU}: the model can perform temporal understanding task, \eg temporal grounding and localization. \textbf{\#Pairs}: the instruction-response pairs for instruction tuning. \textbf{LLM-size}: the large language model adopted in a method.}
\label{tab:qa_results}
\vspace{-1.0em}
\end{table*}
\subsection{Experiment Setups}
\subsubsection{Metrics.}
We evaluate temporal understanding using tasks across various domains: Video Question Answering (Video QA), Audio-visual Video Question Answering (AVQA), and Audio-Visual Event Dense Localization (AVEDL). For General Video QA, zero-shot evaluation is performed on the MSVD-QA~\cite{msvd}, MSRVTT-QA~\cite{msrvtt}, and ActivityNet-QA~\cite{yu2019activitynet} datasets, with open-ended QA tasks evaluated using GPT scoring~\cite{video-chatgpt}. AVQA tasks are assessed on the AVSD~\cite{avsd} and MUSIC-AVQA~\cite{music-avqa} datasets. The AVEDL task uses the UnAV-100~\cite{unav100} dataset, with performance measured by mean Average Precision (mAP) at Intersection over Union (tIoU) thresholds [0.5:0.1:0.9] and the average mAP across [0.1:0.1:0.9].

\subsubsection{Baseline Models.}
For QA tasks, we evaluate LLM-based models, including PandaGPT~\cite{pandagpt}, Macaw-LLM~\cite{macawllm}, and AV-LLM~\cite{avllm}, which support audio-visual input. All evaluated models have 7B or 13B parameters. For AVEDL tasks, we include non-LLM baselines like VSGN~\cite{vsgn}, TadTR~\cite{tadtr}, ActionFormer~\cite{zhang2022actionformer}, UnAV~\cite{unav100}, and UniAV-AT/ST~\cite{geng2024uniav}.

\subsubsection{Implementation Details.}
We uniformly extract a minimum of 100 frames from each video to create an interleaved sequence of audio-visual tokens. Video and audio frames are synchronized through interpolation or downsampling to match the Audio-Interleaving Rate, resulting in a sequence of 100 tokens, each representing 1\% of the audio-visual video duration. The video scaling factor range is set to $[T_{\text{min}}, T_{\text{max}}] = [0.5, 2]$ with a step of 0.1, allowing playback speed adjustments from half to double the original rate. Detailed information on clustering and fine-tuning settings, including specific learning rates and epochs, is provided in the Appendix.

\subsection{Comparison Experiments}
\subsubsection{General Video QA and AVQA.}
The video QA and AVQA comparison results are shown as \Cref{tab:qa_results}. AVicuna supports both audio and video as input and handles temporal understanding tasks. Despite being fine-tuned with only 1.1M pairs and utilizing an LLM with 7B parameters, AVicuna surpasses all other LLM-based models on both video QA and AVQA benchmarks.

\begin{table}[]
\centering
\resizebox{\columnwidth}{!}{%
\setlength{\tabcolsep}{3pt}
\begin{tabular}{l|cccccc}
\toprule
\multicolumn{1}{l|}{\textbf{Method}} & \textbf{0.5} & \textbf{0.6} & \textbf{0.7} & \textbf{0.8} & \textbf{0.9} & \textbf{Avg.} \\ \midrule\midrule
VSGN~\cite{vsgn} & 24.5 & 20.2 & 15.9 & 11.4 & 6.8 & 24.1 \\
TadTR~\cite{tadtr} & 30.4 & 27.1 & 23.3 & 19.4 & 14.3 & 29.4 \\
ActionFormer~\cite{zhang2022actionformer} & 43.5 & 39.4 & 33.4 & 27.3 & 17.9 & 42.2 \\
UnAV~\cite{unav100} & 50.6 & 45.8 & 39.8 & 32.4 & 21.1 & 47.8 \\
UniAV-AT~\cite{geng2024uniav} & 54.1 & 48.6 & 42.1 & 34.3 & 20.5 & 50.7 \\
UniAV-ST~\cite{geng2024uniav} & \multicolumn{1}{l}{54.8} & \multicolumn{1}{l}{49.4} & \multicolumn{1}{l}{43.2} & \multicolumn{1}{l}{35.3} & \multicolumn{1}{l}{22.5} & \multicolumn{1}{l}{51.7} \\
\rowcolor[HTML]{ECF4FF} 
AVicuna (ours) & \textbf{60.0} & \textbf{50.4} & \textbf{49.6} & \textbf{43.5} & \textbf{36.5} & \multicolumn{1}{l}{\cellcolor[HTML]{ECF4FF}\textbf{60.3}} \\ \bottomrule
\end{tabular}%
}
\caption{Comparison of the results on the UnAV-100 for the AVEDL task.}
\label{tab:tu_results}
\vspace{-1.2em}
\end{table}

\begin{table}[!ht]
\centering
\resizebox{\columnwidth}{!}{
\begin{tabular}{l|cccccc}
\toprule
\textbf{Setting}    & \textbf{0.5}             & \textbf{0.6}             & \textbf{0.7}             & \textbf{0.8}            & \textbf{0.9}            & \textbf{Avg.}            \\ \midrule\midrule
\rowcolor[HTML]{ECF4FF}{AVicuna}  & \textbf{60.0}            & \textbf{54.4}            & \textbf{49.6}            & \textbf{43.5}           & \textbf{37.1}           & \textbf{60.3}            \\
w/o PU-VALOR        & \multicolumn{1}{c}{19.5} & \multicolumn{1}{c}{14.3} & \multicolumn{1}{c}{10.2} & \multicolumn{1}{c}{6.8} & \multicolumn{1}{c}{4.5} & \multicolumn{1}{c}{27.9} \\
w/o AVTI            & 50.1                     & 45.2                     & 40.2                     & 34.2                    & 29.4                    & 51.1                     \\
w/o A5-222K         & \multicolumn{1}{c}{22.2} & \multicolumn{1}{c}{16.5} & \multicolumn{1}{c}{11.4} & \multicolumn{1}{c}{6.8} & \multicolumn{1}{c}{2.7} & \multicolumn{1}{c}{30.1} \\
w/o Audio       & 29.0          & 23.9          & 18.8          & 13.6          & 8.8           & 35.8          \\
\bottomrule
\end{tabular}
}
\caption{Ablation study on the dataset and model components, which lead to decreases in mAP.}
\label{tab:abl}
\vspace{-1em}
\end{table}

\subsubsection{Audio-Visual Event Localization.}
The comparison results on the AVEDL tasks can be found in \Cref{tab:tu_results}. In the AVEDL task, AVicuna's superior mAP scores, particularly at the IoU threshold of 0.5 through 0.9, indicate its enhanced precision in localizing events within a video. The results are impressive, considering they outperform other LLM-based models and specialized non-LLM methods like VSGN, TadTR, ActionFormer, and UnAV. This suggests that the AVicuna model has effectively leveraged its audio-visual capabilities to provide a more nuanced understanding of the temporal aspects of videos.
We also conducted the video temporal grounding (VTG) task on the ActivityNet Captions dataset, which can be found in the Appendix.

\subsection{Ablation Study}

\begin{figure}
    \centering
    \includegraphics[width=1.0\linewidth]{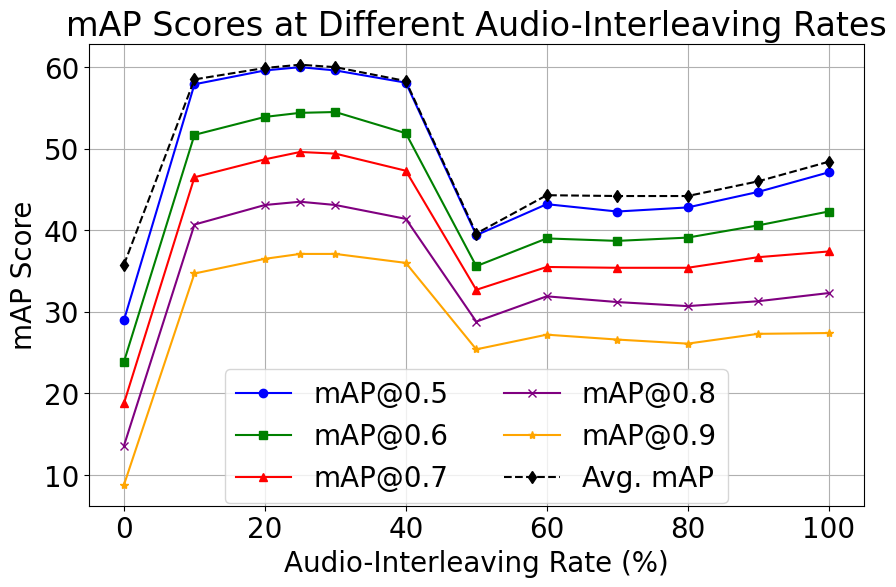}
    \caption{AVicuna's performances on UnAV-100 measured by mAP scores at different AIRs.}
    \label{fig:audio_interleaving_rate}
    \vspace{-1.5em}
\end{figure}

We conduct ablation studies as shown in \Cref{tab:abl} to assess the impact of different components, datasets, and modalities on AVicuna's performance. Each row represents an independent experiment where a specific component or dataset is removed. 
Specifically, omitting the PU-VALOR or A5-222K datasets, especially PU-VALOR, leads to significant performance drops, emphasizing their critical roles in Time-Event Alignment and Audio-Text Alignment, respectively. Removing the AVTI module also results in a decrease in performance, further validating its necessity. Additionally, excluding audio inputs causes a marked reduction in accuracy, underscoring the value of a multimodal approach.

\subsection{Audio-Interleaving Rates Analysis}
The ablation study further sheds light on the contribution of different components to the model's performance. We conduct ablation studies on the Audio-Interleaving Rates (AIR), denoted by $\rho$. \Cref{fig:audio_interleaving_rate} shows AVicuna's performances on UnAV-100 datset for AVEDL task. The scores generally increase with the AIR up to 25-30\%, indicating optimal audio-visual integration, and then decline, suggesting that excessive audio information may be detrimental. The performance gap between different IoU thresholds narrows at higher AIRs. The highest mAP is achieved at 25\% AIR across all thresholds, indicating an optimal balance between audio and video information. The complete results are shown in the Appendix.


\subsection{Qualitative Analysis} 
\Cref{fig:qual} presents three examples of AVicuna's audio-visual dialogue capabilities. In the first video, involving a Jeep with persistent engine noise, AVicuna accurately identifies the time intervals where the engine is knocking. The second video shows people live-streaming while eating pomegranates; AVicuna correctly skips the intro, focuses on the specified time interval, and describes the chewing sounds. In the third video, where two people play silent musical instruments, AVicuna accurately responds with ``NO'', showcasing its ability to provide correct negative responses—a common challenge for existing models. These examples demonstrate AVicuna's flexibility in handling videos of varying lengths and resolutions. More examples can be found in the Appendix.

\begin{figure}[!ht]
    \centering
\includegraphics[width=0.9\linewidth]{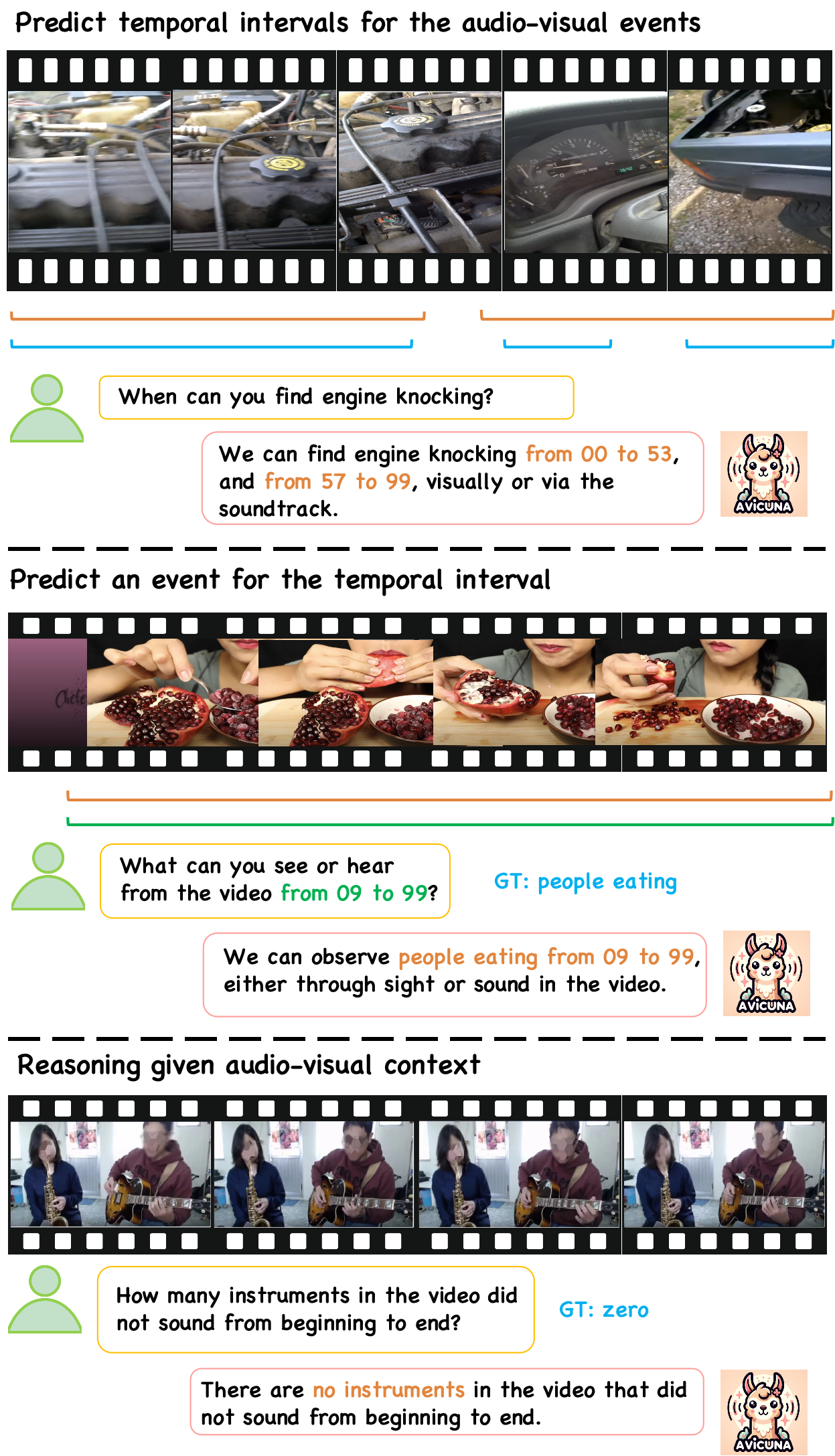}
    \caption{Qualitative results. Blue indicates ground-truth, green indicates the time intervals the user gives, and orange represents the model predictions. AVicuna supports audio-visual video input with various durations and resolutions. Given user queries about an event, it predicts temporal intervals accurately. A given temporal interval provides an accurate response. It also performs reasoning given a question about audio-visual context.}
    \label{fig:qual}
    \vspace{-1.5em}
\end{figure}

\section{Conclusion}
We present a comprehensive approach to enhancing audio-visual temporal understanding in untrimmed videos using MLLMs. We introduced a novel method to construct a pseudo-untrimmed dataset, PU-VALOR, which provides pseudo-untrimmed audio-visual videos with accurate temporal boundary annotations, addressing the scarcity of datasets in the audio-visual domain. Furthermore, we developed AVicuna, an audio-visual LLM incorporating an AVTI and Time-Event Alignment to achieve fine-grained understanding and temporal synchronism in audio-visual videos. Our experiments demonstrate that AVicuna achieves state-of-the-art performance in various video and audio-visual understanding tasks, supporting both coarse-grained QA and fine-grained temporal understanding.

\bigskip

\bibliography{aaai25}
\appendix
\section{Appendix}
\subsection{More Implementation Details}
The dense audio frames are captured using sliding windows of varying sizes and strides. After clustering, 25,270 clusters are produced, with the option to select between 3 and 20 videos for constructing the pseudo-untrimmed video. AVicuna is fine-tuned on a single NVIDIA 48G A6000 GPU. 
Fine-tuning details are as follows: Stage I and II were both fine-tuned for 2 epochs using a learning rate of 1e-3. Stage III and IV were each fine-tuned for 1 epoch using a learning rate of 1e-4. Fine-tuning times were 5/2/36/6 hours for stages I/II/III/IV, respectively.

\subsection{More Comparison Experiments}
\subsubsection{Video-based Generative Performance.}
\Cref{tab:gpt_score} presents the GPT-based evaluation across multiple dimensions for video understanding, as per the methodology of Video-ChatGPT. Our method, AVicuna, achieves the highest scores in temporal understanding (2.53), consistency (2.59), and correctness (2.81) while scoring competitively in detail orientation and context understanding. The superior performance in temporal understanding is particularly notable, attributed to AVicuna's innovative Audio-Visual Tokens Interleaver (AVTI) and Time-Event Alignment. These features enable AVicuna to model long video inputs effectively, capturing the temporal dynamics and ensuring a coherent understanding of the video content over time. The AVTI ensures that audio and visual information is synchronized and integrated, allowing for a more nuanced interpretation of events and their temporal relationships. Additionally, the Time-Event Alignment further enhances the model's ability to discern the temporal boundaries of different events within the video, contributing to its overall temporal understanding, consistency, and correctness.
\begin{table}[!ht]
\centering
\resizebox{\columnwidth}{!}{
\begin{tabular}{l|ccccc}
\toprule
\textbf{Methods} & \textbf{~Temporal~} & \textbf{~Consistency~} & \textbf{~Correctness~} & \textbf{~Detail~} & \textbf{~Context~} \\ \midrule\midrule
VideoChat        & 1.82              & 1.79                 & 2.23                 & 2.50            & 2.53             \\
Video-LLaMA      & 1.98              & 2.15                 & 1.96                 & 2.18            & 2.30             \\
Video-ChatGPT    & 1.98              & 2.37                 & 2.40                 & 2.52            & 2.62             \\
Valley           & 2.04              & 2.45                 & 2.43                 & 2.13            & 2.86             \\
AV-LLM           & 2.17              & 2.51                 & 2.56                 & 2.47            & 2.93             \\
VTimeLLM         & 2.49              & 2.47                 & 2.78                 & \textbf{3.10}   & \textbf{3.40}    \\
\rowcolor[HTML]{ECF4FF}{AVicuna (ours)}   & \textbf{2.53}     & \textbf{2.59}        & \textbf{2.81}        & \underline{2.62}            & \underline{3.25}             \\ \bottomrule
\end{tabular}
}
\caption{GPT-based evaluation for video-based generative performance (1-5) on temporal understanding, consistency, correctness, detail orientation, and context understanding.}
\label{tab:gpt_score}
\end{table}

\subsubsection{Performance on the VTG Task.}
For the Video Temporal Grounding (VTG) task on the ActivityNet Captions dataset, AVicuna's performance in R1@0.5, R1@0.7, and mIoU metrics outstrips that of VideoChat, Video-ChatGPT, and VTimeLLM, reflecting its ability to accurately identify and understand the relevant segments of the video following the textual queries. Though we use official code with all original settings and training steps, our reproduction of VTimeLLM yields slightly lower performance than reported.

\begin{table}[!ht]
\centering
\resizebox{\columnwidth}{!}{%
\begin{tabular}{l|ccc}
\hline
\textbf{Method} & \textbf{R1@0.5} & \textbf{R1@0.7} & \textbf{mIoU} \\ \hline
VideoChat~\cite{li2023videochat} & 3.7 & 1.5 & 7.2 \\
Video-ChatGPT~\cite{video-chatgpt} & 13.6 & 6.1 & 18.9 \\
VTimeLLM~\cite{huang2023vtimellm} & 23.9 & 9.6 & 26.6 \\
\rowcolor[HTML]{ECF4FF}AVicuna (ours) & \textbf{25.0} & \textbf{13.2} & \textbf{28.1} \\ \hline
\end{tabular}%
}
\caption{Comparison of the results on the ActivityNet Captions for VTG task.}
\label{tab:vtg}
\end{table}

\subsection{Dataset Summary}
The datasets and the corresponding question-answering or instruction-response pairs used at each stage are summarized in \Cref{tab:data}.
\begin{table*}[!ht]
\centering
\resizebox{0.98\textwidth}{!}{%
\begin{tabular}{l|c|c}
\toprule
\textbf{Fine-tuning Stages} & \textbf{Data Sources} & \textbf{\#Pairs} \\ \midrule\midrule
Vision-Text Alignment & LCS-558K~\cite{llava} & 558K \\ \hline
Audio-Text Alignment & \begin{tabular}[c]{@{}c@{}}A5-222K~(AudioSet~\cite{audioset}, AudioCap~\cite{kim-etal-2019-audiocaps},\\  Auto-ACD~\cite{autocad})\end{tabular} & 222K \\ \hline
Time-Event Alignment & PU-VALOR (ours), InternVid~\cite{wang2023internvid} & 248K \\ \hline
Instruction Tuning & \begin{tabular}[c]{@{}c@{}}UnAV-100~\cite{unav100}, VideoInstruct100K~\cite{video-chatgpt}, \\ ActivityNet Captions~\cite{anet_cap}, DiDeMo~\cite{didemo}\end{tabular} & 49K \\ \hline
Total & - & 1.1M \\ \bottomrule
\end{tabular}%
}
\caption{Datasets used in multi-stage fine-tuning.}
\label{tab:data}
\end{table*}

\subsection{Prompts and Templates}
We utilize image-text pair prompts from the LCS-588K dataset to generate prompts for Vision-Text Alignment. To ensure symmetry, we transform these image prompts into their audio counterparts for Audio-Text Alignment, as illustrated in Box 1.

\begin{figure}[!ht] 
\centering
\begin{tcolorbox}[colback=white, colframe=gray, text width=0.85\columnwidth, title={\small Box 1: Prompts for Audio-Text Alignment.}, fontupper=\small, fontlower=\small]
$Q_1$: Render a clear and concise summary of the audio.\\
$Q_2$: Write a terse but informative summary of the audio clip.\\
$Q_3$: Present a compact description of the audio's key features.\\
$Q_4$: What is in the audio?\\
$Q_5$: Describe the audio concisely.\\
$Q_6$: Share a concise interpretation of the provided audio.\\
$Q_7$: Give a brief description of the audio.\\
$Q_8$: Provide a brief description of the given audio.\\
$Q_9$: Summarize the auditory content of the audio.
\end{tcolorbox}
\end{figure}

\begin{figure}[!ht] 
\centering
\begin{tcolorbox}[colback=white, colframe=gray, text width=0.85\columnwidth, title={\small Box 2: Templates for adapting labels from AudioSet to responses in A5-222K.}, fontupper=\small, fontlower=\small]
\textbf{Prefix Templates} \\
$R_1$: There is the sound of $\langle$event$\rangle$.\\
$R_2$: I can hear the sound of $\langle$event$\rangle$.\\
$R_3$: Listening to the sound of $\langle$event$\rangle$.\\
$R_4$: Resonating is the sound of $\langle$event$\rangle$.\\
$R_5$: Filling the air is the sound of $\langle$event$\rangle$.\\
$R_6$: There are the sounds of $\langle$event$\rangle$$_1$, $\langle$event$\rangle$$_2$, ...\\
$R_7$: I can hear the sounds of $\langle$event$\rangle$$_1$, $\langle$event$\rangle$$_2$, ...\\
$R_8$: Listening to the sounds of $\langle$event$\rangle$$_1$, $\langle$event$\rangle$$_2$, ...\\
$R_9$: Surrounding me are the sounds of $\langle$event$\rangle$$_1$, $\langle$event$\rangle$$_2$, ...\\
$R_{10}$: Echoing are the sounds of $\langle$event$\rangle$$_1$, $\langle$event$\rangle$$_2$, ...

\hdashrule[0.5ex]{1\textwidth}{0.5pt}{2mm}
\textbf{Suffix Templates} \\
$R_{11}$: $\langle$event$\rangle$ can be heard.\\
$R_{12}$: $\langle$event$\rangle$ is audible.\\
$R_{13}$: $\langle$event$\rangle$ resounds.\\
$R_{14}$: $\langle$event$\rangle$$_1$, $\langle$event$\rangle$$_2$, ... can be heard.\\
$R_{15}$: $\langle$event$\rangle$$_1$, $\langle$event$\rangle$$_2$, ... are audible.\\
$R_{16}$: $\langle$event$\rangle$$_1$, $\langle$event$\rangle$$_2$, ... resound.\\
$R_{17}$: $\langle$event$\rangle$ resounds.\\
$R_{18}$: $\langle$event$\rangle$ permeates the air.\\
$R_{19}$: $\langle$event$\rangle$ is noticeable.\\
$R_{20}$: $\langle$event$\rangle$$_1$, $\langle$event$\rangle$$_2$, ... resound.\\
$R_{21}$: $\langle$event$\rangle$$_1$, $\langle$event$\rangle$$_2$, ... permeate the air.\\
$R_{22}$: $\langle$event$\rangle$$_1$, $\langle$event$\rangle$$_2$, ... are noticeable.
\end{tcolorbox}
\end{figure}

Regarding the A5-222K dataset's audio captions, the audio-text pairs are sourced from the AudioCap and Auto-CAD datasets. However, the AudioSet dataset lacks audio captions and instead provides labels. We utilize templates to convert these labels into responses, presented in Box 2.

\subsection{Completed AIR Results}
The completed AIR (audio-interleaving rate) results are shown in \Cref{tab:air_complete}.
\begin{table}
\centering
\scalebox{1}{
\begin{tabular}{l|cccccc}
\toprule
\textbf{$\rho$} & \textbf{0.5}       & \textbf{0.6}       & \textbf{0.7}       & \textbf{0.8}       & \textbf{0.9}       & \textbf{Avg.}          \\ \midrule\midrule
0\%       & 29.0          & 23.9          & 18.8          & 13.6          & 8.8           & 35.8          \\
10\%                    & 57.9          & 51.7          & 46.5          & 40.7          & 34.7          & 58.5          \\
20\%                    & 59.6          & 53.9          & 48.7          & 43.1          & 36.5          & 59.9          \\
\rowcolor[HTML]{ECF4FF}{25\%}                    & \textbf{60.0} & 54.4          & \textbf{49.6} & \textbf{43.5} & \textbf{37.1} & \textbf{60.3} \\
30\%                    & 59.6          & \textbf{54.5} & 49.4          & 43.1          & \textbf{37.1} & 60.0          \\
40\%                    & 58.1          & 51.9          & 47.3          & 41.4          & 36.0          & 58.3          \\
50\%                    & 39.4          & 35.6          & 32.7          & 28.8          & 25.4          & 39.6          \\
60\%                    & 43.2          & 39.0          & 35.5          & 31.9          & 27.2          & 44.3          \\
70\%                    & 42.3          & 38.7          & 35.4          & 31.2          & 26.6          & 44.2          \\
80\%                    & 42.8          & 39.1          & 35.4          & 30.7          & 26.1          & 44.2          \\
90\%                    & 44.7          & 40.6          & 36.7          & 31.3          & 27.3          & 46.0          \\
100\%      & 47.1          & 42.3          & 37.4          & 32.3          & 27.4          & 48.4          \\ \bottomrule
\end{tabular}
}
\caption{The impact of AIRs on mAP at different IoU thresholds on the UnAV-100 dataset
for the AVEDL task.}
\label{tab:air_complete}
\end{table}

\subsection{More Qualitative Results}
\begin{figure*}[]
    \centering
\includegraphics[width=0.9\linewidth]{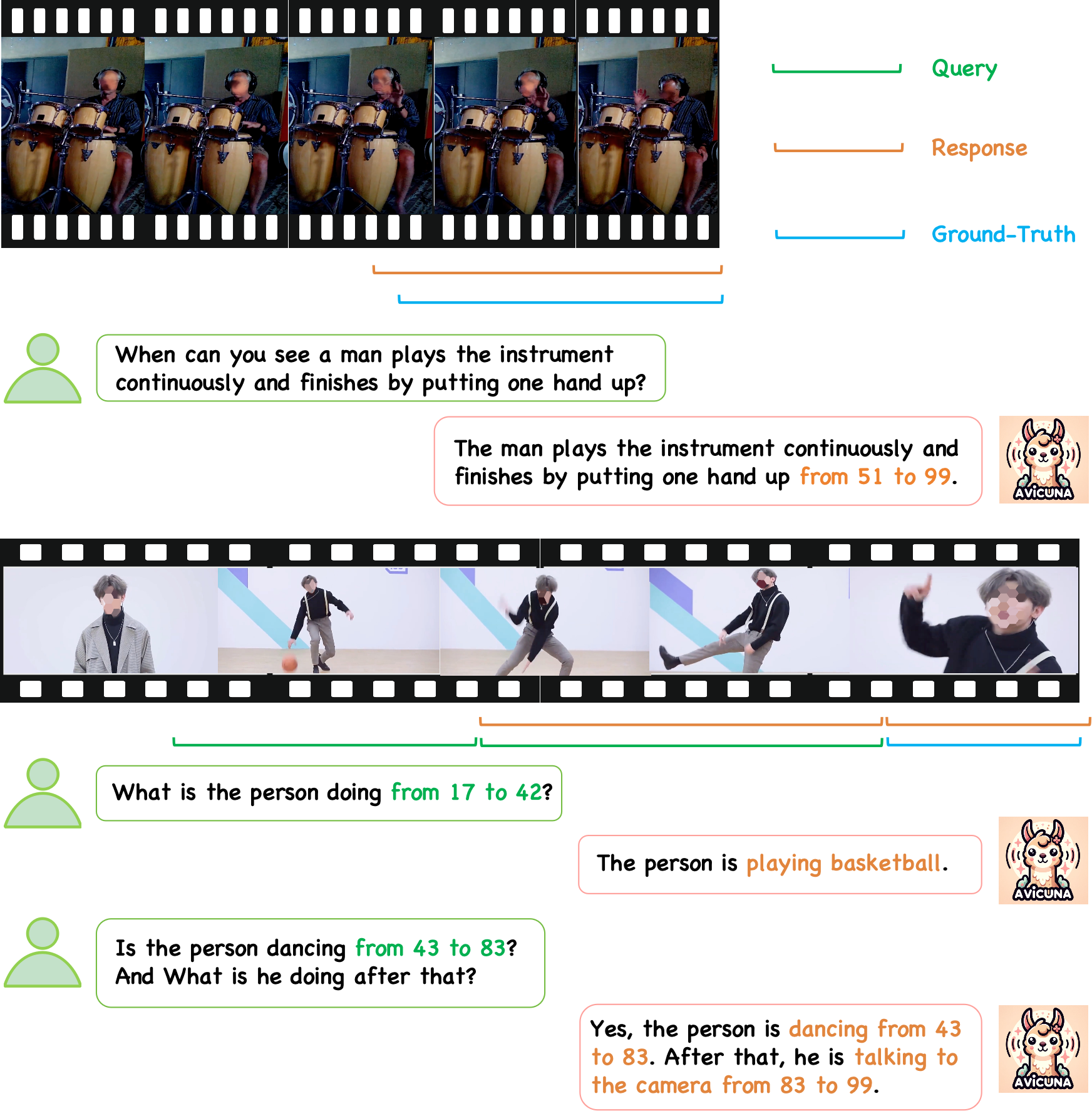}
    \caption{More results. This figure illustrates AVicuna's capability to localize events by interpreting natural language queries and offering precise event descriptions based on provided timestamps. Additionally, AVicuna can process complex queries to provide accurate and context-aware responses, showcasing its superior performance in video temporal understanding.}
    \label{fig:qual1}
\end{figure*}

\begin{figure*}[]
    \centering
\includegraphics[width=0.9\linewidth]{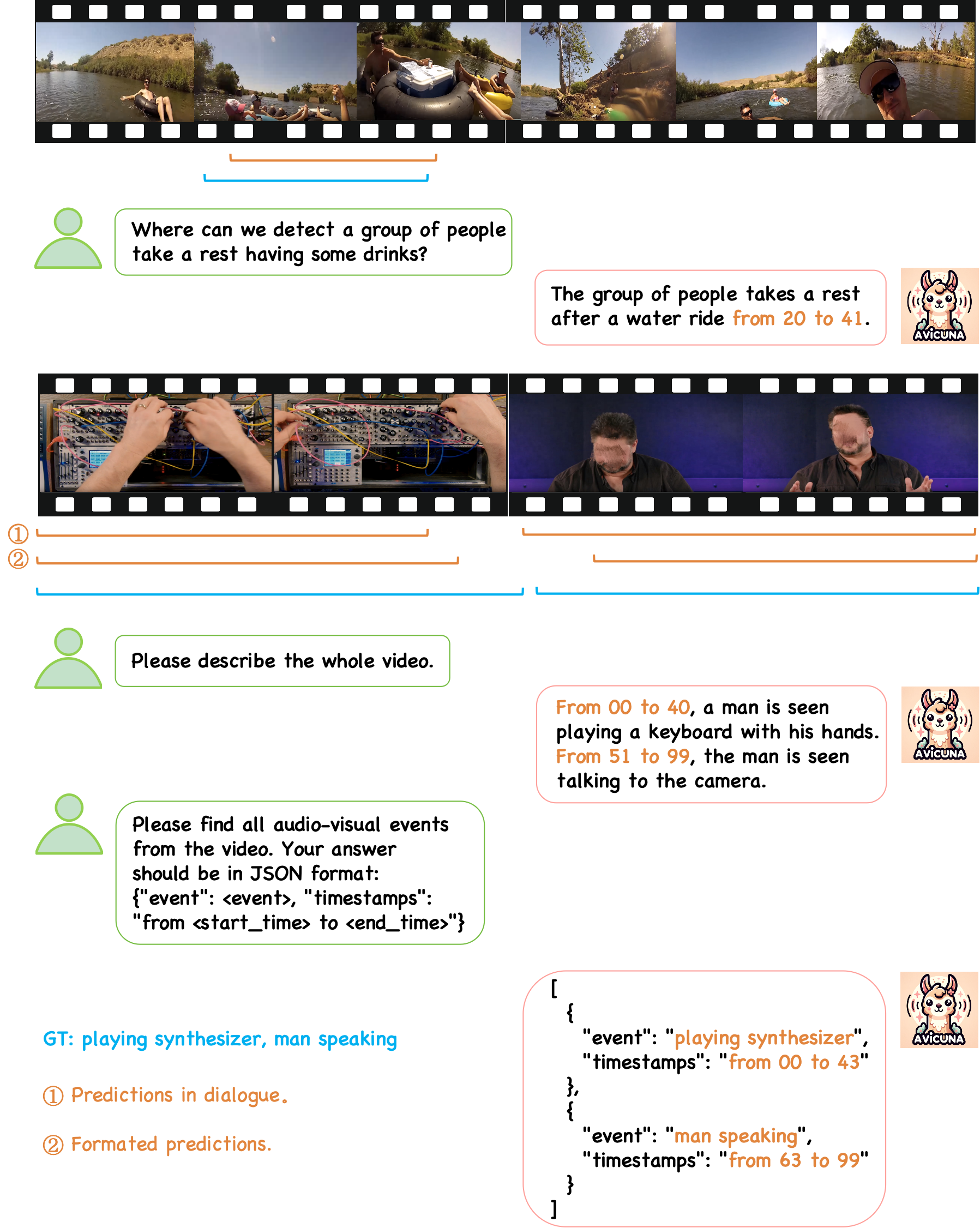}
    \caption{More results (continue). AVicuna's advanced functionality allows it to provide detailed video captions that include pairs of timestamps and narratives, going beyond basic captioning. By utilizing LLMs, AVicuna can output responses in JSON format for audio-visual event dense localization, facilitating easy parsing and integration with downstream models. }
    \label{fig:qual2}
\end{figure*}

\begin{figure*}[]
    \centering
\includegraphics[width=0.9\linewidth]{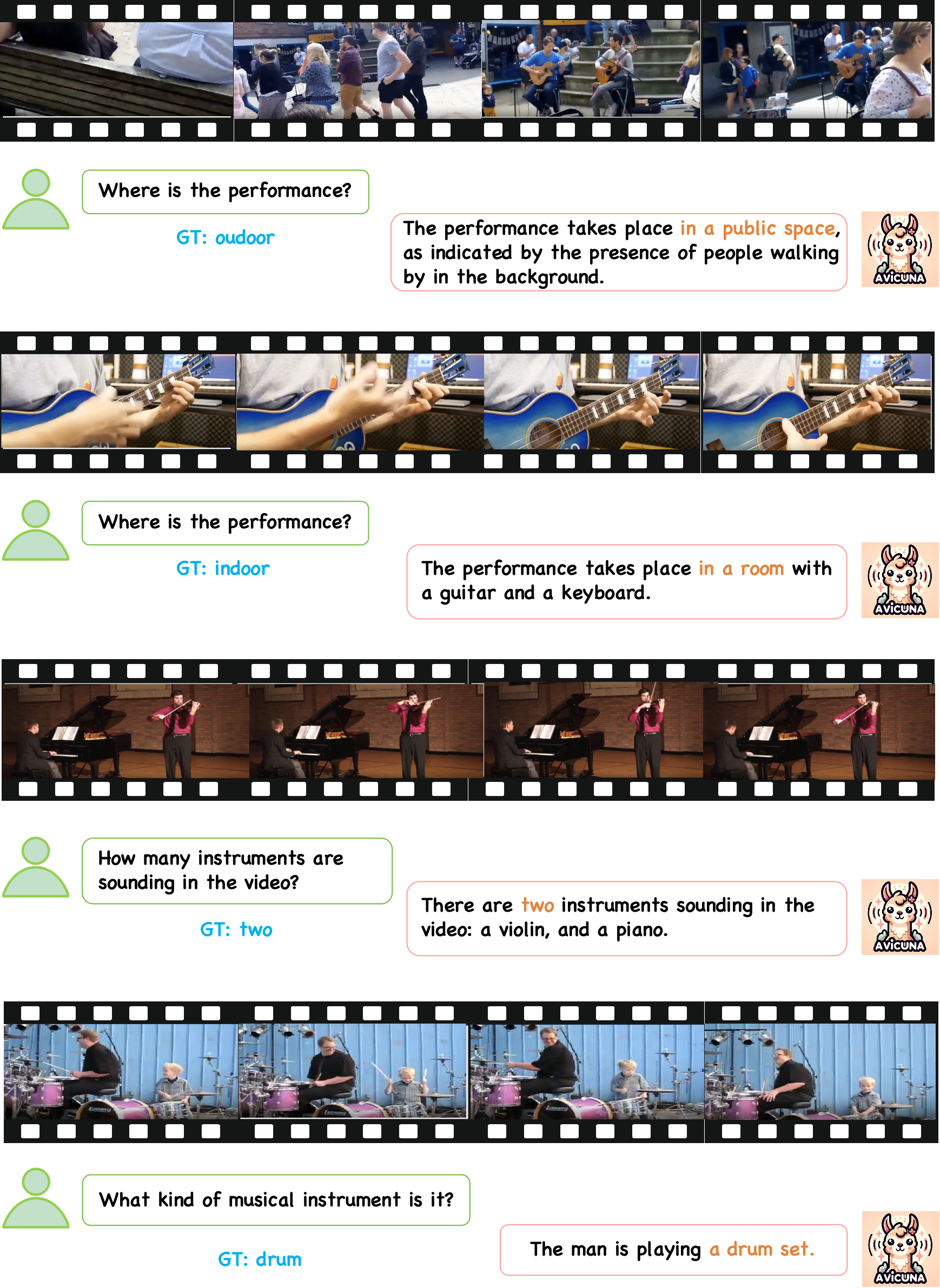}
    \caption{More results (continue). AVicuna can provide detailed answers for enhanced audio-visual video QA, from identifying objects and scenes to accurate counting.}
    \label{fig:qual3}
\end{figure*}

Here, we provide more qualitative results to show the AVicuna's performance, which are shown in \Cref{fig:qual1,fig:qual2,fig:qual3}.

\subsection{Limitations}
Even though AVicuna achieves good performance in a variety of fine-grained video understanding tasks and referential dialogue, there are still some limitations.
\subsubsection{Hallucination.} One notable challenge is hallucination, where the model generates plausible but incorrect details not present in the input data. This can lead to misinformation if not adequately addressed and may affect the model's reliability in critical applications.
\subsubsection{Deficiency in Spatial Comprehension.} While AVicuna provides advancements in temporal understanding, spatial-temporal grounding in long-form or untrimmed videos remains an area for improvement. The model’s ability to simultaneously localize and understand events spatially and temporally in a video is still not at par with human comprehension.
\subsubsection{Insufficient Precision for Ultra-Long Videos.} The precision of representing video time percentages using 100 natural language-formatted numbers is limited, especially for ultra-long videos. This can impact the model's ability to accurately localize and understand events in videos with extended durations.

\end{document}